\documentclass[runningheads]{llncs}
\usepackage{graphicx}
\usepackage{hyperref}
\usepackage{algorithm, algorithmic}
\usepackage{amsmath, amssymb}
\begin{document}

\title{A Data-Driven Probabilistic Framework for Cascading Urban Risk Analysis Using Bayesian Networks}
\author{Chunduru Rohith Kumar\inst{1, 2} \and
PHD Surya Shanmuk\inst{1, 3} \and
Prabhala Naga Srinivas\inst{1, 4} \and
Sri Venkatesh Lankalapalli\inst{1, 5} \and
Debasis Dwibedy\inst{1, 6}
}

\authorrunning{Chunduru et al.}
\institute{Department of Computer Science and Engineering, School of Computing, Coimbatore, Amrita Vishwa Vidyapeetham, Tamil Nadu, India, 641112 \\ \and
\email{cb.en.u4cse21612@cb.students.amrita.edu}\\ \and
\email{cb.en.u4cse21646@cb.students.amrita.edu}\\ \and
\email{cb.en.u4cse21647@cb.students.amrita.edu} \\ \and
\email{cb.en.u4cse21660@cb.students.amrita.edu}\\ \and
\email{d$\_$debasis@cb.amrita.edu}\\
}

\maketitle              
\begin{abstract}
 
The increasing complexity of cascading risks in urban systems necessitates robust, data-driven frameworks to model interdependencies across multiple domains. This study presents a foundational Bayesian network–based approach for analyzing cross-domain risk propagation across key urban domains including air, water, electricity, agriculture, health, infrastructure, weather, and climate. Directed Acyclic Graphs (DAGs) are constructed using Bayesian Belief Networks (BBNs), with structure learning guided by Hill-Climbing search optimized through Bayesian Information Criterion (BIC) and K2 scoring. The framework is trained on a hybrid dataset that combines real-world urban indicators with synthetically generated data from Generative Adversarial Networks (GANs), and is further balanced using the Synthetic Minority Over-sampling Technique (SMOTE). Conditional Probability Tables (CPTs) derived from the learned structures enable interpretable probabilistic reasoning and quantify the likelihood of cascading failures. The results identify key intra and inter-domain risk factors and demonstrate the framework’s utility for proactive urban resilience planning. This work establishes a scalable, interpretable foundation for cascading risk assessment and serves as a basis for future empirical research in this emerging interdisciplinary field.
\end{abstract}

\section{Introduction}\label{sec: Introduction}
The accelerated development of urban systems, augmented by integrated digital technologies, aims to enhance infrastructure, service delivery, and quality of life [2]. However, this interconnectivity also introduces systemic vulnerabilities. In particular, cascading failures, where a disruption in one subsystem triggers a chain reaction across other dependent systems, have emerged as a significant concern [3, 4]. For instance, a power outage can impair water distribution, disrupt emergency medical services, and compound existing environmental stressors, leading to wide-scale urban dysfunction [2].\\
\textbf{Research Motivation.} Traditional risk assessment models often operate in silos, focusing on domain-specific threats without accounting for the interdependencies between urban subsystems [8]. Such limitations hinder proactive risk management, especially in complex, hyper-connected ecosystems where secondary effects can be more disruptive than the primary incident itself [1, 5]. As cities become smarter and more interconnected, the vulnerability to these multi-domain cascading risks becomes more pronounced.\\
\textbf{Our Contribution.} To address the above-mentioned challenges, we introduce a novel data-driven probabilistic framework for modeling and analysing cascading risk events across urban city domains using Bayesian Belief Networks (BBNs). We employ structure learning algorithms, specifically the Hill-Climbing method optimized via Bayesian Information Criterion (BIC) and K2 scoring to construct Directed Acyclic Graphs (DAGs) that capture causal dependencies among risk factors. The datasets used include real-world urban indicators from sources such as NDAP and OpenCity, which are further augmented through synthetically generated records using Generative Adversarial Networks (GANs). To handle the data imbalance issue and improve coverage, we incorporate the Synthetic Minority Over-sampling Technique (SMOTE). The resulting Conditional Probability Tables (CPTs) allow interpretable estimation of cascading failures across domains such as water, air, electricity, agriculture, health, and infrastructure. Our approach is distinguished by its ability to integrate real and synthetic data, uncover both intra and inter-domain dependencies, and provide quantified, interpretable risk estimates that support proactive urban planning. These insights are directly applicable in energy management, healthcare, disaster preparedness, and infrastructure resilience. Overall, we aim to offer a scalable and empirically validated tool for system-wide risk inference in urban cities, enabling decision-makers to anticipate and mitigate the impact of cascading failures.\\
\textbf{Paper Organization.} The remaining sections of the paper is organized as follows. In Section 2, we present key definitions and relevant background, including a review of prior work on Bayesian Networks and cascading risk modeling. Section 3 formally defines the problem statement. Section 4 details the proposed methodology, including data acquisition, synthetic data generation using GANs, data preprocessing, network structure learning, and probabilistic inference through Conditional Probability Tables. In Section 5, we present the experimental results, graphical representations of inter and intra-domain risk propagation, and interpretations of key findings. We conclude the paper in Section 6 with a summary of the main contributions and a discussion on potential future research directions.
\section{Background Studies} \label{sec: Background Studies}
\subsection{Basic Terminologies, Notations, and Definitions} \label{subsec:Basic Terminologies, Notations, and Definitions}
Here, we define key terms, definitions and notations frequently used throughout the paper.
\begin{itemize}
\item \textbf{Risk Factor (RF):} A quantitative metric that represents the likelihood of an adverse event occurring within a specific domain. For example, in the context of urban cities, critical risk factors include air quality index, traffic density, and the status of water resources [3].

\item \textbf{Cascading Events:} A sequence of events where an initial disruption in one domain triggers subsequent failures in interconnected systems. In urban cities, for example, a pollution spike can impact public health, or traffic congestion may elevate emission levels [1].

\item \textbf{Domain (D):} Distinct domains of a urban city such as transportation, healthcare, natural resources (sub-domains: water, air, electricity, agriculture), and climate. Each domain is evaluated for risk exposure and its potential to influence other domains through cascading effects [2].

\item \textbf{Generative Adversarial Network (GAN):} A machine learning framework used to create synthetic datasets that reflect the statistical patterns of real data. In this study, GANs are utilized to simulate domain-specific data, such as correlations between air pollution levels and health outcomes [10].

\item \textbf{Synthetic Minority Over-sampling Technique (SMOTE):} A re-sampling technique applied to imbalanced datasets to generate synthetic data for minority classes. In our methodology, SMOTE is used to balance the dataset before integrating it with GAN-generated synthetic data [5].

\item \textbf{Probability of Cascading Events (PCE):} A probabilistic metric that estimates the likelihood of a cascading failure occurring in a specific domain due to the influence of a primary risk factor. This metric forms the foundation for predicting chain reactions across systems in urban cities [18].
\end{itemize}
\subsection{Literature Survey} \label{subsec:Literature Survey}
Bayesian Networks (BNs) have been widely adopted in recent years to model complex dependencies among variables, particularly in fields where uncertainty and interrelationships are critical. Song et al. [16] applied the Max-Min Hill-Climbing (MMHC) algorithm to construct BNs for analyzing multimorbidity using longitudinal health data from over 19,000 individuals. Their study demonstrated that BNs could effectively capture both direct and indirect causal dependencies among variables such as sleep duration, age, and physical activity, offering advantages over traditional logistic regression models in revealing complex interrelationships. Although their focus was limited to clinical decision-making and statistical association, the methodological insight into structure learning with CPTs and probabilistic inference is pertinent to risk modeling across domains.\\
Complementing this, Adhitama et al. [17] used a score-based Hill-Climbing algorithm, optimized via Bayesian Information Criterion (BIC) to model dependencies among 52 symptoms and 15 eye diseases. The resulting BN comprising 93 edges and 65 nodes enabled probabilistic diagnostic reasoning and achieved a minimal BIC score after iterative refinement. This study further reinforces the efficacy of structure learning techniques, such as BIC scoring, in constructing interpretable and domain-specific probabilistic models.
While these works primarily address healthcare-centric or domain-restricted problems, they provide a foundation for extending BNs to broader applications, such as multi-domain cascading risks in urban cities. Our work builds on these methodologies by integrating cross-domain real and synthetic datasets, and adapting BN-based modeling to represent interdependencies across heterogeneous urban systems like water, air, electricity, and health. Unlike the cited studies, which focus on single-domain scenarios, we contribute a generalized and scalable probabilistic framework tailored to the complexity of interconnected urban infrastructures.
Although Bayesian Networks have been applied in various domains, there is a notable scarcity of research explicitly addressing data-driven cascading risk analysis in interconnected urban systems. Most existing works either examine isolated domains or employ expert-driven frameworks with limited scalability. We believe this work to be among the initial attempts at developing an integrated probabilistic model using real and synthetic data sources to infer cascading risks across multiple urban domains. We thus aim to set a methodological foundation for future advancements in this emerging interdisciplinary area.
\section{Problem Statement} \label{subsec: Problem Statement}
In urban cities, domains such as air quality, water availability, energy infrastructure, public health, and agriculture are deeply interconnected. A disruption in one domain can propagate to others, resulting in cascading failures. The challenge is to model these interdependencies in a data-driven and interpretable manner, such that potential risk propagation paths can be identified and quantified in advance. Formally, we define the problem as follows:


%
\begin{itemize}
\item
\textbf{Input:} Multi-modal real and synthetic datasets from Urban city domains including Air, Water, Electricity, Agriculture, Weather, Climate, Health, and Infrastructure. We denote the datasets as follows:
\[
\mathbf{D} = \{D_{\text{Air}}, D_{\text{Water}}, D_{\text{Electricity}}, D_{\text{Agriculture}}, D_{\text{Weather\_Climate}}, D_{\text{Health}}, D_{\text{Infrastructure}}\}
\]
\item \textbf{Expected Output:} Probabilistic estimates of cascading risk propagation pathways across domains, represented through interpretable structures such as Directed Acyclic Graphs (DAGs) and Conditional Probability Tables (CPTs).
\item \textbf{Objective:} To develop a data-driven probabilistic framework that can learn cross-domain dependencies and infer the likelihood of cascading failures, enabling early warning and proactive risk management in urban cities.
\end{itemize}

\section{Our Proposed Framework } \label{sec: Our Proposed Framework }

%

The various phases of our proposed framework is depicted in Figure 1.
    \begin{figure}[h!]
    \centering
    \includegraphics[width=1.10\textwidth]{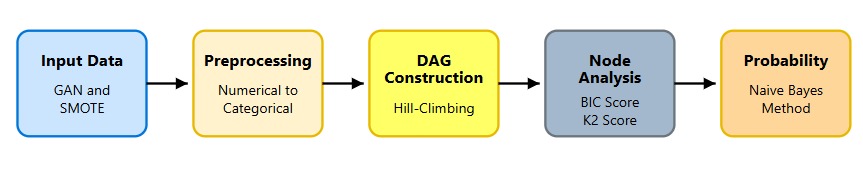}
    \caption{Phases of the Proposed Cascading Risk Modeling Framework}
    \label{fig:workflow}
\end{figure}
\begin{itemize}
\item \textbf{Data Collection and Dataset Generation.}
We consider eight critical domains such as air, water, electricity, agriculture, health, infrastructure, weather and climate and for each domain we identify relevant risk indicators. We collect 10 data tuples for every domain from official government sources such as National Data and Analytics Platform (NDAP) and OpenCity. To enrich the dataset and improve its statistical coverage, we employ a Conditional Generative Adversarial Network (cGAN) to generate synthetic data that preserves the domain-specific distributions and interdependencies. To handle the class imbalance issues in the collected data, we apply the Synthetic Minority Oversampling Technique (SMOTE), ensuring balanced representation across high-risk and low-risk instances. Our final dataset captures both intra and inter-domain variations and is publicly available at: https://github.com/Srinivas-2004/Cascading-Risk-Prediction.
\item \textbf{Data Preprocessing.} To facilitate binary risk modeling, we convert all numerical features into categorical risk indicators, assigning 1 to high-risk values and 0 to low-risk ones. This binarization enhances interpretability and supports clear threshold-based intervention analysis.
\item \textbf{Structure Learning and DAG Construction.} We use a Bayesian Belief Network (BBN) to model causal dependencies among domain-specific attributes. A Directed Acyclic Graph (DAG) is constructed using the Hill-Climbing algorithm, a score-based structure learning approach. The algorithm iteratively refines the graph to maximize a scoring criterion, starting from an initial configuration and performing local modifications until convergence.
\begin{itemize}
\item \textbf{Node-Relationship Analysis.} We evaluate candidate graph structures using two well-known scoring functions such as  Bayesian Information Criterion (BIC) Score and K2 Score. The functions are formally defined below. 
\begin{center}
\begin{equation}
    BIC = -2*ln(L)+k*ln(n)
    \end{equation}
\end{center}
where L denote the model likelihood, k denote the number of parameters, and n denote the number of data samples.\\ Lower BIC values indicate better models.\\\\
\textbf{K2 Score:}
\begin{center}
\begin{equation}
    K2 = \prod_{i=1}^{N} \prod_{j=1}^{q_i} 
    \frac{(N_{ij} - 1)!}{\prod_{k=1}^{r_i} N_{ijk}!} 
    \cdot \frac{(r_i - 1)!}{(N_{ij} + r_i - 1)!}
\end{equation}
\end{center}
where, G denote the Bayesian network structure (graph), D denote the dataset, n  is the number of nodes (variables), $q_i$ denote the number of unique parent configurations for node $i$, and $r_i$ is the number of possible values of node $i$, $N_{ij}$ denote the number of cases in the dataset where the parents of node $i$ take the $j$-th configuration, $N_{ijk}$ is the number of cases where node $i$ takes its $k$-th value while its parents take the $j$-th configuration. \\
Edges are retained based on score improvements: lower BIC or higher K2 scores indicate stronger dependencies.\\
\end{itemize}
\item \textbf{Risk Probability Estimation.} From the final learned structure, we derive Conditional Probability Tables (CPTs) for each target attribute (e.g., overall risk level) based on its parent nodes in the DAG. These tables enable probabilistic reasoning and simulate cascading effects by computing:
\begin{center}
    \begin{equation}
    P(C \mid X) = \frac{P(X \mid C) P(C)}{P(X)}
\end{equation}
\end{center}
where, $P(C \mid X)$ denote the posterior probability of class C given the features X, $P(X \mid C)$ denote the likelihood, i.e., the probability of the features X given class C, $P(C)$ is the prior probability of class C, $P(X)$ is the evidence, i.e., the overall probability of features X occurring. 
\end{itemize}
\section{Results and Discussion} \label{sec: Results and Discussion}
This section presents the results of our Bayesian Network–based risk modeling for eight interconnected urban domains. Using the learned DAG structures, we identify both intra and inter-domain dependencies and interpret the Conditional Probability Tables (CPTs) to estimate cascading risks.
\subsection{DAG-Based Dependency Structure Analysis}
Directed Acyclic Graphs (DAGs) is constructed for each domain using BIC and K2 scoring. For analysis, we selected the DAG from each domain that exhibited the highest number of statistically meaningful dependencies. Figures 2–4 depict domain-specific and cross-domain DAGs. These visualizations reveal significant causal patterns.
\begin{itemize}
\item In natural resource domains with sub-domains such as water, air, electricity, agriculture, strong intra-domain links are observed, e.g., air pollutants such as NO2 and VOCs influence each other and ultimately affect air quality risk.
\item Infrastructure and health domains exhibit fewer but stronger causal edges, reflecting the high impact of fewer critical factors such as ICU capacity or death rate.
\item Cross-domain DAGs (Fig. 4) underscore how weather variables like atmospheric pressure and temperature impact water availability and public health indicators, forming pathways of cascading failure.
\end{itemize}
\begin{figure}[H]
    \centering
    \begin{minipage}{0.48\textwidth}
        \centering
        \fbox{\includegraphics[width=\linewidth]{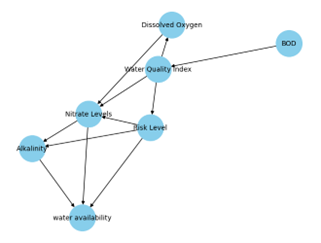}} \\
        (a) Water
    \end{minipage}
    \hfill
    \begin{minipage}{0.48\textwidth}
        \centering
        \fbox{\includegraphics[width=\linewidth]{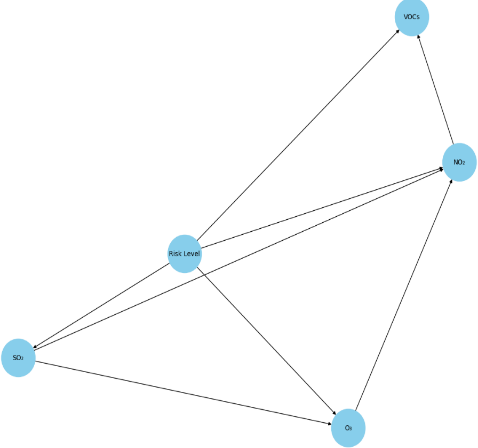}} \\
        (b) Air
    \end{minipage}
    \vspace{0.5cm}
    \begin{minipage}{0.48\textwidth}
        \centering
        \fbox{\includegraphics[width=\linewidth]{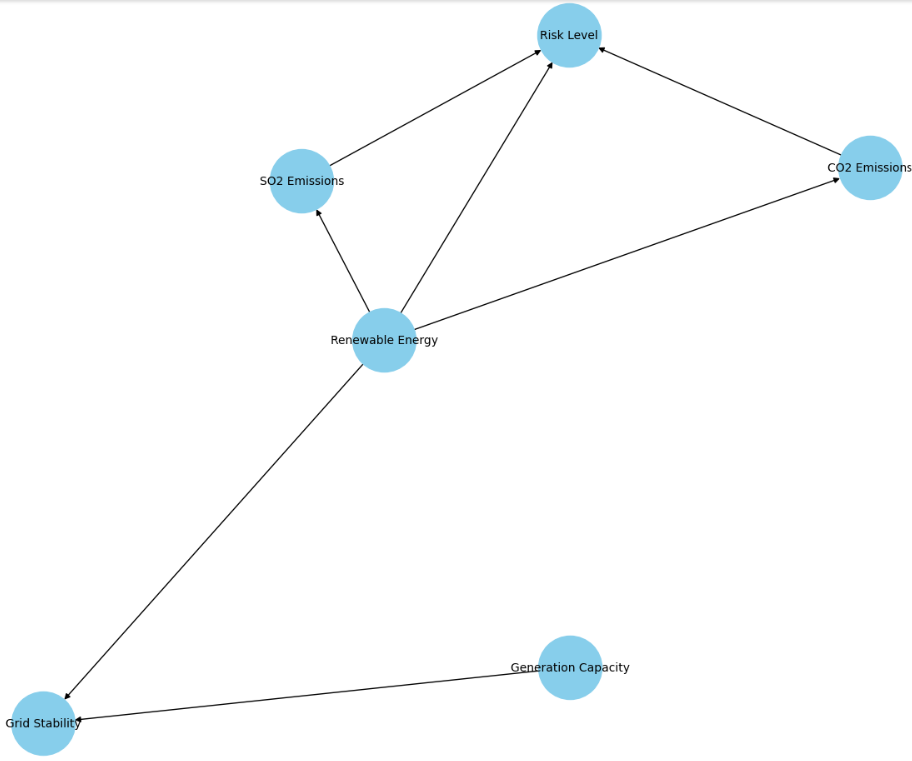}} \\
        (c) Electricity
    \end{minipage}
    \hfill
    \begin{minipage}{0.48\textwidth}
        \centering
        \fbox{\includegraphics[width=\linewidth]{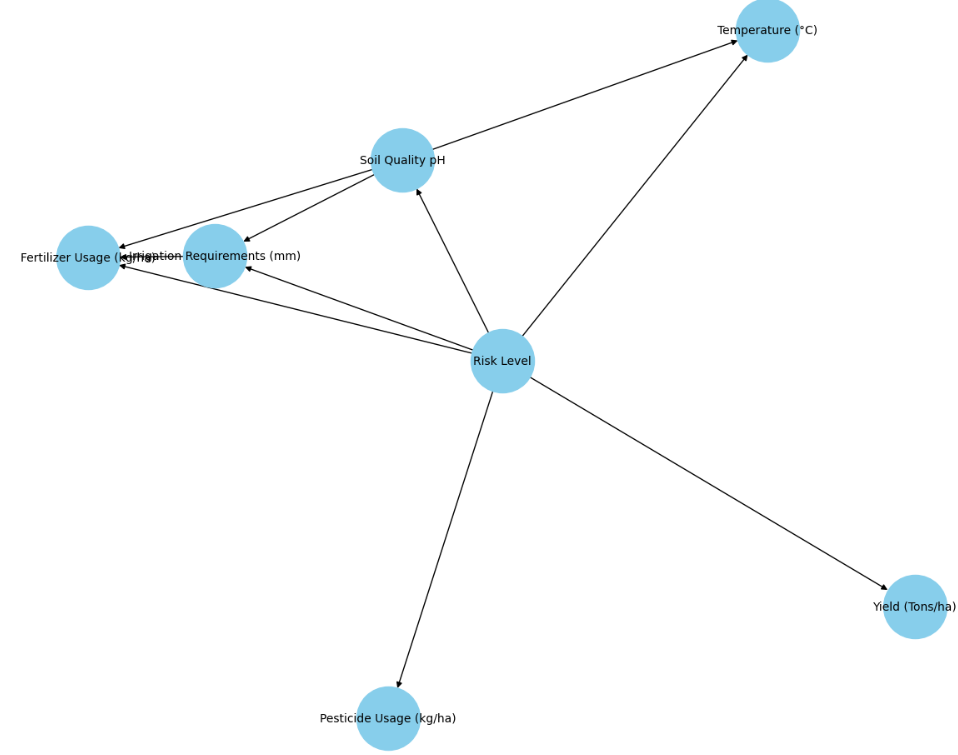}} \\
        (d) Agriculture
    \end{minipage}
    \caption{DAGs for Natural resources sub-domians (a) Water (b) Air (c) Electricity (d) Agriculture}
    \label{fig:DAG}
\end{figure}
\begin{figure}[H]
    \centering
    \begin{minipage}{0.23\textwidth}
        \centering
        \fbox{\includegraphics[width=\linewidth]{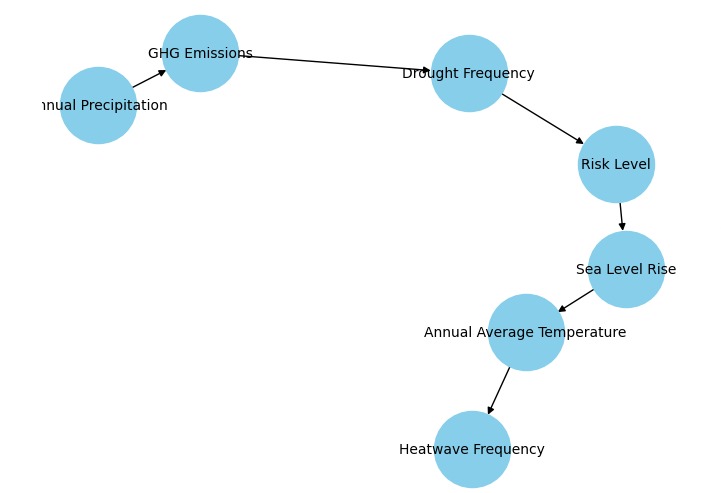}} \\
        (a) Climate
    \end{minipage}
    \hfill
    \begin{minipage}{0.23\textwidth}
        \centering
        \fbox{\includegraphics[width=\linewidth]{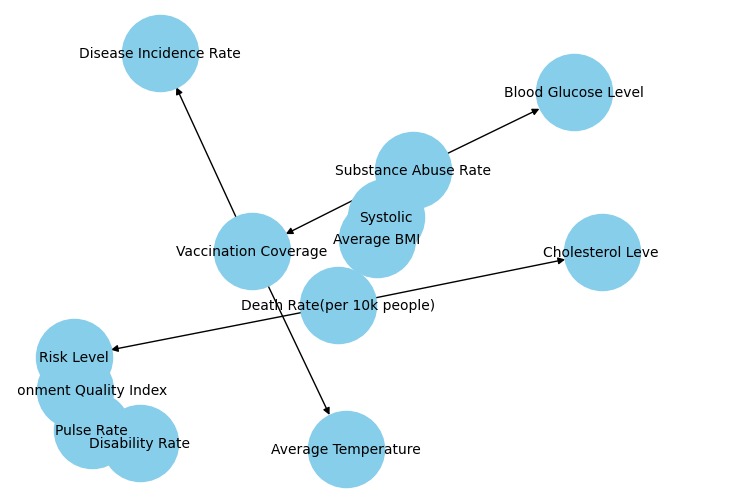}} \\
        (b) Weather
    \end{minipage}
    \vspace{0.5cm}
    \begin{minipage}{0.23\textwidth}
        \centering
        \fbox{\includegraphics[width=\linewidth]{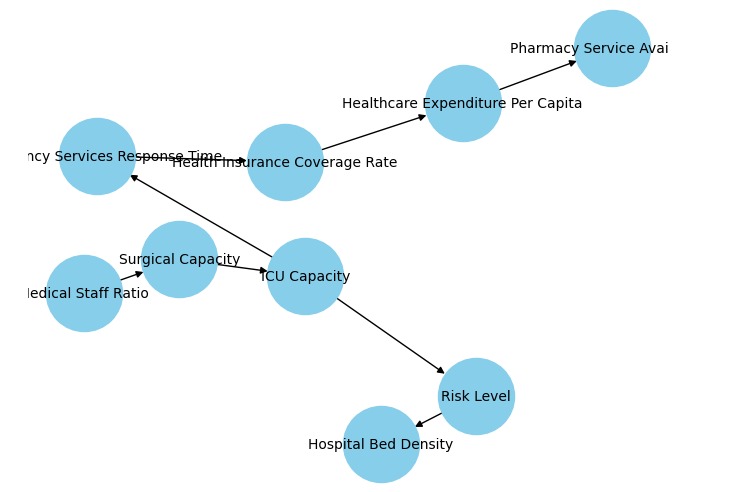}} \\
        (c) Infrastructure
    \end{minipage}
    \hfill
    \begin{minipage}{0.23\textwidth}
        \centering
        \fbox{\includegraphics[width=\linewidth]{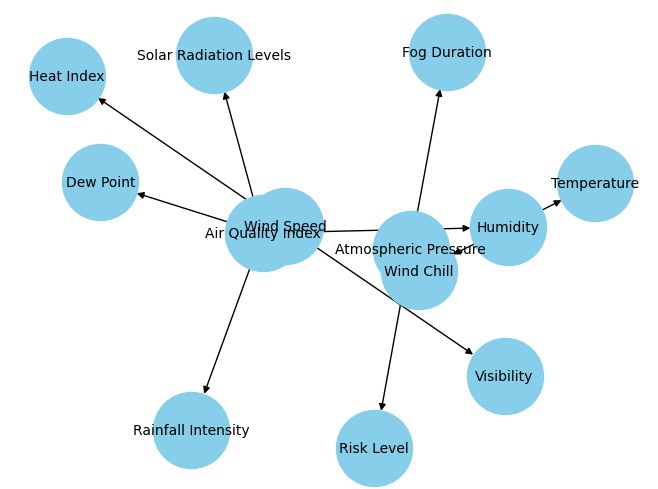}} \\
        (d) Health
    \end{minipage}
    \caption{DAGs of Domains (a) Climate, (b) Health, (c) Infrastructure, (d) Weather}
    \label{fig:DAG}
\end{figure}
\begin{figure}[H]
    \centering
    \begin{minipage}{0.45\textwidth}
        \centering
        \fbox{\includegraphics[width=1.1\linewidth]{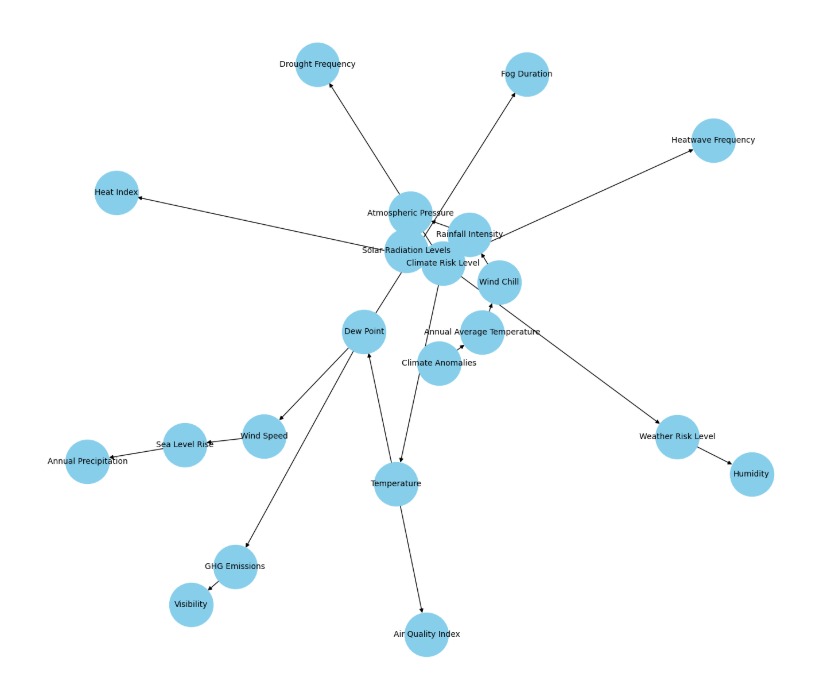}} \\
        (a) Weather and Climate
    \end{minipage}
    \hfill
    \begin{minipage}{0.45\textwidth}
        \centering
        \fbox{\includegraphics[width=1.1\linewidth]{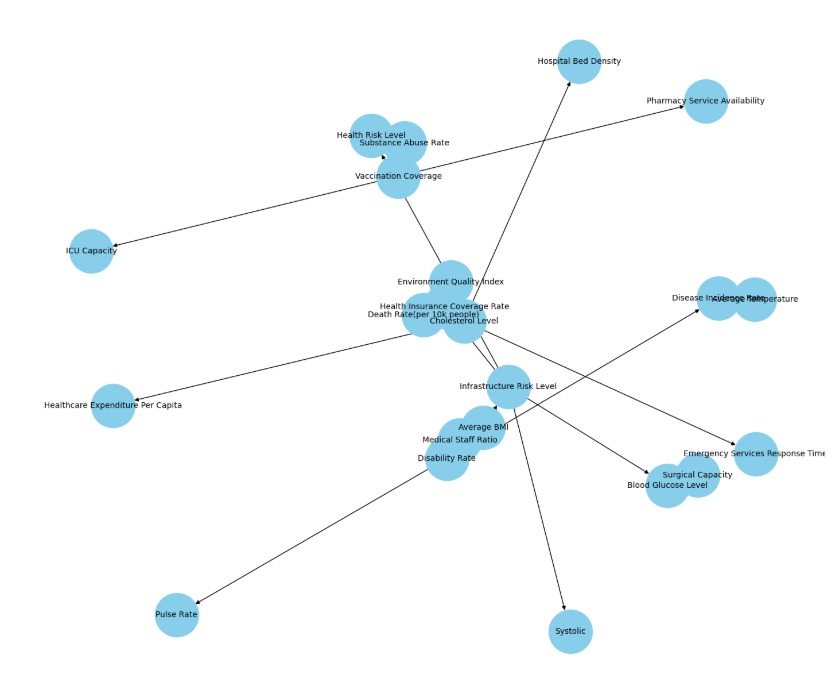}} \\
        (b) Health and Infrastructure
    \end{minipage}
    \vspace{0.5cm}
    \begin{minipage}{0.45\textwidth}
        \centering
        \fbox{\includegraphics[width=1.1\linewidth]{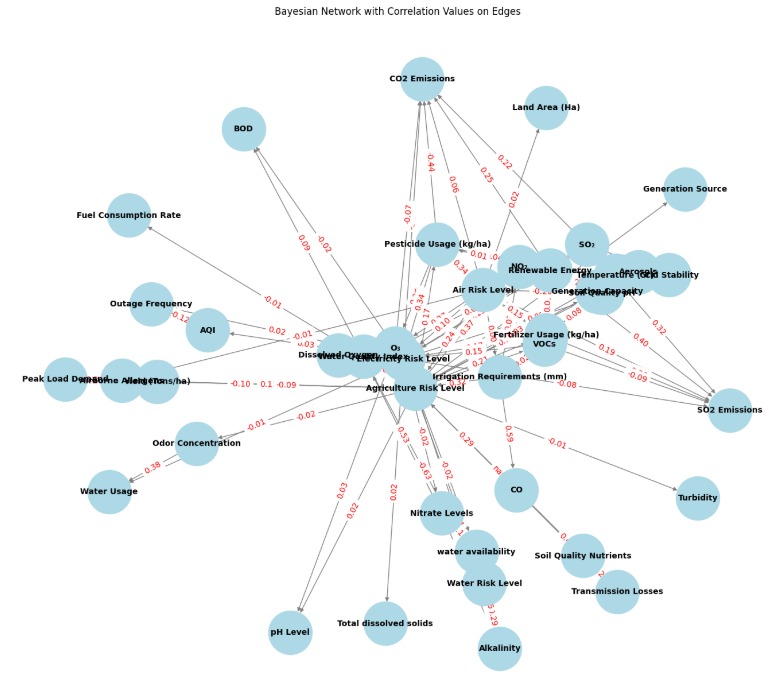}} \\
        (c) Natural Resources
    \end{minipage}
    \hfill
    \begin{minipage}{0.45\textwidth}
        \centering
        \fbox{\includegraphics[width=1.1\linewidth]{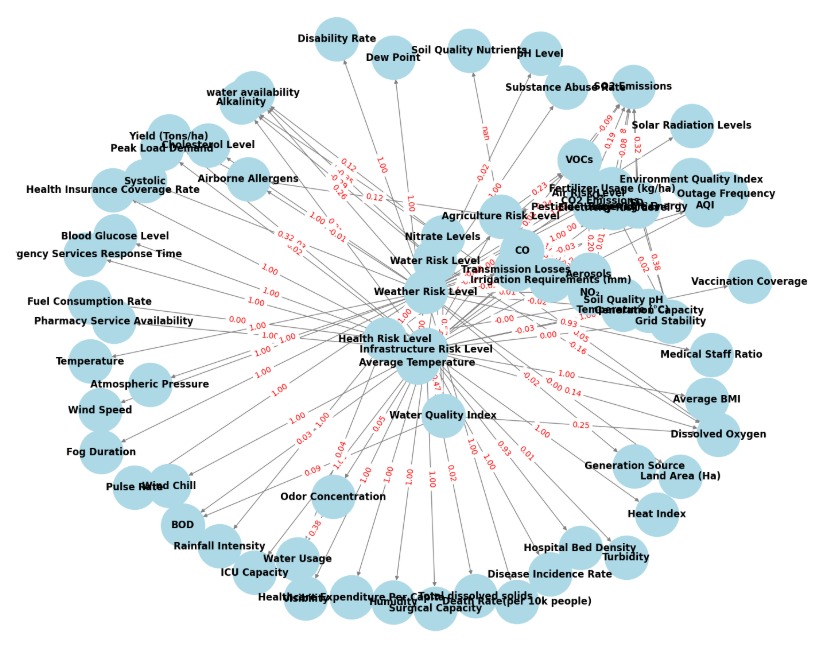}} \\
        (d) All domains
    \end{minipage}
    \caption{DAGs of domains (a) Weather and Climate, (b) Health Infrastructure, (c) Natural Resources, (d) All Domains}
    \label{fig:DAG}
\end{figure}
\subsection{Domain-Level Risk Estimation}
For each domain, we compute CPTs to estimate the posterior probability of a domain entering a high-risk state given the values of its parent attributes. Key results are summarized below.\\\\
\textbf{Water:} The Table 1 represents the CPT for the water sub-domain.
\begin{table}[H]
    \centering
    \begin{tabular}{|l|c|c|c|c|}
        \hline
        \textbf{Water Quality Index} & \textbf{Water Quality Index(0)} & \textbf{Water Quality Index(1)} \\
        \hline
        Risk Level(0)  & $0.0564$ & $0.7221$ \\
        Risk Level(1) & $0.9435$ & $0.2778$ \\
        \hline
    \end{tabular}
    \caption{CPT for Water}
    \label{tab:performance_metrics}
\end{table}
\textit{Inference:} The Water Quality Index (WQI) alone has a dominant influence. A low WQI increased the probability of a high-risk state from $5.6\%$ to $94.3\%$, illustrating its criticality in water safety management.\\\\
\textbf{Air:} The Table 2 represents the CPT for the air sub-domain.
\begin{table}[H]
    \centering
    \begin{tabular}{|l|c|c|c|c|}
        \hline
        \textbf{NO2} & \textbf{NO2(0)} & \textbf{NO2(1)} & \textbf{NO2(1)} \\
        \hline
        \textbf{O3} & \textbf{O3(0)} & \textbf{O3(1)} & \textbf{O3(1)} \\
        \hline
        \textbf{SO2} & \textbf{SO2(0)} & \textbf{SO2(1)} & \textbf{SO2(1)} \\
        \hline
        \textbf{VOC} & \textbf{VOC(0)} & \textbf{VOC(1)} & \textbf{VOC(1)} \\
        \hline
        Risk Level(0)  & $0.9980$ & $0.2168$ & $0.1698$ \\
        Risk Level(1) & $0.0019$ & $0.7831$ & $0.8301$ \\
        \hline
    \end{tabular}
    \caption{CPT for Air}
    \label{tab:performance_metrics}
\end{table}
\textit{Inference:} Pollutants including NO2, O3, SO2, and VOCs contributed to high-risk states. When all four pollutants are at elevated levels, the probability of high-risk air conditions exceeds to $83\%$, highlighting a compounded effect.\\\\
\textbf{Electricity:} The Table 3 represents the CPT for the electricity sub-domain.
\begin{table}[H]
    \centering
    \begin{tabular}{|l|c|c|c|c|}
        \hline
        \textbf{CO2 Emissions} & \textbf{CO2 Emissions(0)} & \textbf{CO2 Emissions(1)} & \textbf{CO2 Emissions(1)} \\
        \hline
        \textbf{Renewable Energy} & \textbf{Renewable Energy(0)} & \textbf{Renewable Energy(1)} & \textbf{Renewable Energy(1)} \\
        \hline
        \textbf{SO2 Emissions} & \textbf{SO2 Emissions(0)} & \textbf{SO2 Emissions(1)} & \textbf{SO2 Emissions(1)} \\
        \hline
        Risk Level(0)  & $0.6877$ & $0.8912$ & $0.9646$ \\
        Risk Level(1) & $0.3122$ & $0.1087$ & $0.035$ \\
        \hline
    \end{tabular}
    \caption{CPT for Electricity}
    \label{tab:performance_metrics}
\end{table}
\textit{Inference:} High renewable energy usage and low CO2 emissions significantly reduce electricity-related risks. This supports energy policy emphasis on sustainable grid components.\\\\
\textbf{Agriculture:} The Table 4 represents the CPT for the agriculture sub-domain.
\begin{table}[H]
    \centering
    \begin{tabular}{|l|c|c|c|c|}
        \hline
        \textbf{Soil Quality pH} & \textbf{Soil Quality pH(0)} & \textbf{Soil Quality pH(1)}  \\
        \hline
        Risk Level(0)  & 0.975 & 0.024 \\
        Risk Level(1) & 0.894 & 0.105 \\
        \hline
    \end{tabular}
    \caption{CPT for Agriculture}
    \label{tab:performance_metrics}
\end{table}
\textit{Inference:} Soil pH shows a binary effect on agricultural risk. A low pH level increases the high-risk probability to $89.4\%$, emphasizing the need for monitoring agricultural soil health.\\\\
\textbf{Health:} The Table 5 represents the CPT for the health domain.
\begin{table}[H]
    \centering
    \begin{tabular}{|l|c|c|}
        \hline
        \textbf{Death Rate (per 10k people)} & \textbf{Death Rate(0)} & \textbf{Death Rate(1)} \\
        \hline
        Risk Level(0)  & $0.9991680532445923$ & $0.0008319467554076539$ \\
        Risk Level(1)  & $0.0008319467554076539$ & $0.9991680532445923$ \\
        \hline
    \end{tabular}
    \caption{CPT for Health}
    \label{tab:health_cpt}
\end{table}
\textit{Inference:} The only factor that directly affects the Risk Level in this dataset is the Death Rate per 10,000 people. The table shows the probability that the risk level is very high when the death rate is also high. Conversely, when the death rate is low, the probability of a high-risk level significantly decreases.\\\\
\textbf{Infrastructure:} The Table 6 represents the CPT for the hospital infrastructure domain.
\begin{table}[H]
    \centering
    \begin{tabular}{|l|c|c|}
        \hline
        \textbf{ICU Capacity} & \textbf{ICU Capacity(0)} & \textbf{ICU Capacity(1)} \\
        \hline
        Risk Level(0)  & 0.9991680532445923 & 0.0008319467554076539 \\
        Risk Level(1)  & 0.0008319467554076539 & 0.9991680532445923 \\
        \hline
    \end{tabular}
    \caption{CPT for Infrastructure}
    \label{tab:infrastructure_cpt}
\end{table}
\textit{Inference:} The only factor that directly affects the Risk Level in this dataset is ICU Capacity. The table shows that the probability of the risk level being very high is when ICU Capacity is very low. Conversely, when ICU Capacity is high, the probability of a high-risk level significantly decreases, indicating the critical role of healthcare infrastructure in mitigating risk.\\\\
\textbf{Weather:} The Table 7 represents the CPT for the weather domain.
\begin{table}[H]
    \centering
    \begin{tabular}{|l|c|c|}
        \hline
        \textbf{Atmospheric Pressure} & \textbf{Atmospheric Pressure(0)} & \textbf{Atmospheric Pressure(1)} \\
        \hline
        Risk Level(0)  & 0.9991680532445923 & 0.0008319467554076539 \\
        Risk Level(1)  & 0.0008319467554076539 & 0.9991680532445923 \\
        \hline
    \end{tabular}
    \caption{CPT for Weather}
    \label{tab:weather_cpt}
\end{table}
\textit{Inference:} The only factor that directly affects the Risk Level in this dataset is Atmospheric Pressure. The table shows that the probability of the risk level being very high increases when atmospheric pressure is at extreme levels. Conversely, when atmospheric pressure remains within a stable range, the probability of a high-risk level significantly decreases.\\\\
\textbf{Climate:} The Table 8 represents the CPT for the climate domain.
\begin{table}[H]
    \centering
    \begin{tabular}{|l|c|c|}
        \hline
        \textbf{Drought Frequency} & \textbf{Drought Frequency(0.0)} & \textbf{Drought Frequency(1.0)} \\
        \hline
        Risk Level(0)  & 0.9991680532445923 & 0.0008319467554076539 \\
        Risk Level(1)  & 0.0008319467554076539 & 0.9991680532445923 \\
        \hline
    \end{tabular}
    \caption{CPT for Climate}
    \label{tab:climate_cpt}
\end{table}
\textit{Inference:} The only factor that directly affects the Risk Level in this dataset is Drought Frequency. The table shows the probability that the risk level is very high when the drought frequency is also high. Conversely, when the drought frequency is low, the probability of a high-risk level significantly decreases.
\subsection{Cross-Domain Cascading Risk Patterns} \label{subsec: Cross-Domain Cascading Risk Patterns}
The DAGs and CPTs are further analyzed to reveal multi-hop cascading effects across domains. Notable observations are outlined below.
\begin{itemize}
\item In the Water sub-domain, dissolved oxygen and nitrate levels show a strong conditional influence on overall water risk. The CPTs indicate that low dissolved oxygen significantly increases the probability of a high-risk state in water quality.
\begin{itemize}
\item \textbf{Water → Infrastructure → Health:} High Water Usage → Reduced Turbidity Management → Health Risk Level Increase
\end{itemize}
\item For the Air domain, pollutants like NO2, VOCs and SO2 emerged as key drivers. The CPTs show that even marginal increases in these pollutants lead to a higher likelihood of air quality deterioration, confirming their role in urban air toxicity.
\begin{itemize}
\item \textbf{Air → Water → Health:} High VOCs → Low Water Quality Index → High BOD → Public Health Risk \\
In the Electricity domain, renewable energy percentage and CO2 emissions were highly influential. CPTs highlight that low renewable energy penetration, combined with high CO2 levels, leads to elevated electricity risk levels and unstable grid conditions.
\end{itemize}
\item In the health and infrastructure domain, the CPTs expose multi-variable dependencies, where factors like ICU capacity, staff density, and emergency service response times play significant roles in determining the health risk level. Conditional dependencies show that if even one of these factors is in a low state, the risk probability sharply increases.
\begin{itemize}
\item \textbf{Agriculture → Water → Health:} High VOCs → Low Water Quality Index → High BOD → Public Health Risk
\end{itemize}
\end{itemize}
In the Weather and Climate domain, the Conditional Probability Tables (CPTs) highlight temperature, humidity, rainfall, atmospheric pressure, and sea level rise as key influencing factors. Weather-related risk is predominantly driven by anomalies in humidity and temperature, with even minor deviations significantly increasing the likelihood of a high-risk state. Climate risk, on the other hand, is shaped by long-term environmental indicators such as solar radiation, atmospheric pressure fluctuations, and rising sea levels. The CPTs underscore that while each of these factors may individually exert a moderate influence, their combined and cumulative variations can collectively escalate the overall risk, illustrating the compounding nature of environmental stress in cascading failure scenarios.\\
In the Cross-Domain analysis, the Conditional Probability Tables (CPTs) reveal clear patterns of risk propagation across interconnected urban systems. For instance, weather-related stressors are found to significantly impact water availability, while deteriorating air quality increased the likelihood of adverse health outcomes. Similarly, vulnerabilities in infrastructure, such as limited ICU capacity or delayed emergency responses, exacerbates risks in the health sector. A key insight from the CPTs is that cascading failures are more likely to emerge when multiple domains experience moderate stress simultaneously, rather than from a singular extreme event in one domain. This highlights the importance of a holistic, system-wide perspective in risk management, where the combined effect of multiple stressors is considered more critical than isolated domain failures. These insights can guide practical implementations in energy grid management, public health preparedness, and infrastructure planning by simulating possible chains of failure.
\subsection{Discussion} \label{subsec: Discussion}

This study demonstrates how integrating real-world urban indicators with synthetically generated data using Conditional GANs, enhanced through SMOTE-based balancing, allows for robust modeling of cascading risks in urban cities. By employing Bayesian Belief Networks trained via score-based DAG learning (BIC and K2), the proposed framework successfully captures both intra and inter-domain dependencies. The learned structures and corresponding CPTs not only quantify the likelihood of risk escalation but also offer interpretable insights into complex, multi-variable interactions.

A key insight from the results is that cascading failures are more likely to arise from moderate stress across multiple domains rather than from isolated extreme events. This supports a systems-oriented approach to urban resilience, where the compounded influence of domain interconnectivity is accounted for in both analysis and planning. The framework thus provides a transparent and scalable decision-support tool that can guide proactive interventions across infrastructure, public health, energy, and environmental systems.

\section{Conclusion} \label{Conclusion and Future Scope}

This work presents a foundational data-driven probabilistic framework for cascading risk analysis in smart city domains using Bayesian Belief Networks and Directed Acyclic Graphs. By combining real and synthetically generated urban datasets and addressing data imbalance through SMOTE, the framework captures key interdependencies across domains such as air, water, agriculture, electricity, health, and infrastructure. Structure learning via score-optimized DAG construction and probabilistic inference using CPTs enables interpretable estimation of risk propagation across interconnected domains.

In future work, this framework can be extended to incorporate temporal dimensions for real-time forecasting, integrate multi-modal sensor streams for high-resolution monitoring, and embed adaptive decision-making for automated policy response. Techniques such as Kullback–Leibler (KL) divergence can further be employed to assess model fidelity by comparing learned structures with ground-truth baselines. Overall, this framework establishes a scalable, interpretable foundation for cascading risk modeling and supports the development of resilient, data-informed urban management strategies.

\end{document}